\documentclass[manuscript,screen]{acmart}

\usepackage{graphicx}
\usepackage{booktabs}       
\usepackage{tabularx}       
\usepackage{microtype}      

\AtBeginDocument{%
  }

\setcopyright{acmlicensed}

\begin{document}

\title[A Few Good Clauses]{A Few Good Clauses: Comparing LLMs vs Domain-Trained Small Language Models on Structured Contract Extraction}

\author{Nicole Lincoln}
\email{nicole.lincoln@onit.com}

\author{Nick Whitehouse}
\email{nick.whitehouse@onit.com}

\author{Jaron Mar}
\email{jaron.mar@onit.com}

\author{Rivindu Perera}
\email{rivindu.perera@onit.com}
\affiliation{%
  \institution{Onit AI Labs, Onit Inc.}
  \city{Auckland}
  \country{New Zealand}
}

\renewcommand{\shortauthors}{Lincoln et al.}

\begin{abstract}
This paper evaluates whether a domain-trained Small Language Model (SLM) can outperform frontier Large Language Models on structured contract extraction while operating at dramatically lower cost. We test Olava Extract, a self-hosted legal-domain Mixture-of-Experts model, against five commercially available frontier systems on 24 held-out public contracts annotated by legal professionals, covering 508 labelled field instances across 26 extraction targets.
 
Olava Extract achieved the strongest aggregate performance in the study, with a macro F1 of 0.812 and a micro F1 of 0.842, outperforming all evaluated frontier baselines while operating at a fraction of their inference cost. Under batched inference, extraction cost fell by 78\% to 97\% compared with the frontier models tested. Olava Extract also achieved the highest precision scores in the evaluation, producing fewer unsupported extractions and fewer hallucinated answers than the frontier baselines. This distinction is particularly important in legal workflows, where hallucinated outputs can create significant downstream review burden, operational risk, and loss of trust in automated systems.

The implications extend beyond contract extraction itself. For legal services, the results suggest that high-performing, human-comparable legal AI can now operate at infrastructure-scale throughput and materially lower cost using smaller domain-trained systems, increasing the economic pressure on traditional review models built around Junior Lawyers and Legal Process Outsourcers. More broadly, the findings challenge a central assumption of the current AI market: that commercially valuable enterprise AI capability must remain tightly coupled to ever-larger models, massive infrastructure expenditure, and centrally hosted providers. On a commercially meaningful enterprise task, a relatively compact specialised model was able to compete directly with frontier-scale systems while dramatically outperforming them on cost and deployment efficiency.
\end{abstract}

\ccsdesc[500]{Computing methodologies~Natural language generation}
\ccsdesc[500]{Computing methodologies~Information extraction}
\ccsdesc[500]{Applied computing~Law}
\ccsdesc[300]{Applied computing~Document searching}

\keywords{contract extraction, small language models, fine-tuning, legal NLP, self-hosted AI, information extraction}
\maketitle
\section{Introduction}
\label{sec:intro}

Extracting key terms from contracts is one of the most time-consuming tasks in legal operations. Agreements vary widely in structure, terminology, drafting style. Relevant information may appear in clauses, schedules, exhibits, or definitions spread across many pages. Reviewers must identify provisions, dates, parties, obligations, and financial terms while preserving enough source context for the extracted answer to be verified.

Recent work has shown that Large Language Models (LLMs) can automate parts of this work. In \textit{Better Call GPT}~\cite{martin2024}, we evaluated frontier LLMs against Junior Lawyers and Legal Process Outsourcers (LPOs) on contract review. The models matched or exceeded human accuracy while reducing review time and cost. In \textit{Better Bill GPT}~\cite{whitehouse2025}, we extended the comparison to legal invoice review and found that LLMs outperformed human reviewers across the evaluated decision and classification tasks.

Those results show that frontier LLMs can perform structured legal analysis tasks well. They also leave open an important deployment question. The models evaluated in those studies were large, commercially hosted systems accessed through third-party APIs. For legal teams, that setup can create practical constraints around data control, security review, cost predictability, model versioning, and long-term availability. 

Small Language Models (SLMs) offer a new path. They can be fine-tuned on domain-specific data and deployed within an organisation's own infrastructure. If such a model can meet frontier-model quality on contract extraction, it may offer a better fit for workflows where confidentiality, cost control, and deployment stability matter. In this paper, we use "SLM" to include efficient Mixture-of-Experts (MoE) models whose active inference footprint is in the range commonly associated with smaller dense models, even when their nominal parameter count is higher.

This study evaluates Olava Extract, a domain-trained contract extraction model. We test it on a full-document structured extraction task covering 26 common contract fields across 24 held-out public contracts, with 508 human-labelled field instances. The fields include extracted clauses, duration fields, dates, currency values, Boolean and categorical selections, and short text fields identifying information that is either directly stated or inferred from the contract. We compare Olava Extract with five frontier models using the same contract set, output schema, prompt structure, inference approach, and scoring rules.

The study focuses on three questions: 
\begin{enumerate}
    \item Can a domain-trained Small Language Model reach aggregate extraction performance comparable to leading frontier models? 
    \item Are Small Language Models cheaper than Large Language Models, beyond the obvious advantages of self-hosting?
    \item Are Small Language Models faster than Large Language Models?
\end{enumerate}

Together, these questions extend the research direction established in \textit{Better Call GPT} and \textit{Better Bill GPT}. In \textit{Better Call GPT}, we posited that LLMs would disrupt Junior Lawyers and LPOs by showing that they could perform legal review tasks with comparable accuracy, greater speed, and lower cost. This study continues that experimentation by examining a high-volume task: structured contract extraction. The focus now shifts from model capability to production viability. Frontier LLMs can perform legal tasks with strong accuracy, but at the scale required for structured contract extraction, their cost, latency, and data-control constraints may reduce their practical advantage over human review. 

This paper evaluates whether a smaller, domain-trained model can retain sufficient extraction accuracy while providing a more cost-effective and operationally practical path to scaled legal AI deployment.

\section{Related Work}
\label{sec:related}

This study builds on three strands of prior work: legal-domain model adaptation, contract-specific extraction benchmarks, and efficient fine-tuning of smaller models.

Prior legal NLP research has shown that legal text benefits from domain-specific modelling. \citeauthor{chalkidis2020} evaluated strategies for adapting BERT to legal corpora and released a family of legal-domain BERT models for legal NLP applications~\cite{chalkidis2020}. This supports the core premise that legal language is specialised enough to justify domain adaptation rather than relying only on general-purpose models.

Contract review has also become a useful benchmark for legal AI because it requires both structured information extraction with long, variable documents. CUAD introduced an expert-annotated contract-review dataset, framing review as the identification of contract provisions that matter to human reviewers~\cite{hendrycks2021}. ContractNLI approached contract analysis as document-level natural language inference, requiring models to classify hypotheses as entailed, contradicted, or not mentioned and to identify supporting evidence spans~\cite{koreeda2021}. Together, these datasets show that contract analysis is automatable but remains difficult, especially where documents are long, drafting conventions vary, or evidence must be gathered across multiple clauses.

Recent legal-industry evaluations have shown that frontier LLMs can perform structured legal review tasks at or above human-reviewer baselines. Prior studies in this series found strong LLM performance on procurement-contract issue spotting and legal-invoice review, with large gains in speed and cost efficiency~\cite{martin2024,whitehouse2025}. Those studies establish the usefulness of LLMs for structured legal workflows, but they rely on large, API-hosted frontier models.

The present study focuses on whether similar utility can be achieved with a smaller, self-hostable model. Recent work outside law shows that compact models trained on carefully curated data can rival much larger systems on standard benchmarks, as demonstrated by Phi-3~\cite{abdin2024}. In the legal domain, SaulLM-7B shows that continued training on legal corpora can produce capable legal-domain models at modest scale~\cite{colombo2024}. Parameter-efficient fine-tuning methods such as LoRA further reduce the cost of adapting pretrained models by training only a small number of additional parameters~\cite{hu2022}. Sparse Mixture-of-Experts architectures provide another route to efficiency by increasing total model capacity while activating only part of the model for each token~\cite{fedus2022,jiang2024}.

This study extends that line of work by testing Olava Extract, a domain-trained SLM, on full-document structured contract extraction across 26 field types. Unlike CUAD's clause-span benchmark or ContractNLI's inference task, this evaluation measures field-level extraction quality, cost, and deployment practicality. The question is therefore not only whether contract extraction can be automated, but whether a smaller, domain-trained model can make automation economically viable for high-effort legal tasks that have traditionally been assigned to Junior Lawyers, LPOs, and other human reviewers.

\section{Methodology}
\label{sec:methodology}

\subsection{Task Definition and Output Schema}
\label{sec:task}

This study evaluated contract extraction as a full-document structured generation task. Each contract was inserted into the model context without chunking, retrieval, or vector-store augmentation. For evaluation, each model extracted all target fields from each contract in a single-turn LLM call, returning outputs using the same structured schema for Olava Extract and all frontier baselines. This design avoids introducing retrieval, chunking, or multi-step orchestration effects that could confound the comparison. As a result, performance differences more directly reflect each model's ability to interpret the full contract and populate the target schema under the same task conditions.

For each target field, the output schema required two elements: a display answer and supporting verbatim contract span(s). The display answer represents the normalised value or concise answer intended for downstream use. The supporting span(s) provide the source contract text used to justify the answer, consistent with contract-analysis benchmarks that use annotated clause spans or evidence spans, including CUAD and ContractNLI~\cite{hendrycks2021,koreeda2021}. They also allow field-level scoring against human annotations.

The extraction task covered 26 fields across six field categories, shown in Table~\ref{tab:fieldcat}. The categories are used both to describe the task and to interpret field-level performance in the results.

\begin{table*}[ht]
  \centering  
  \small
  \begin{tabularx}{\textwidth}{@{} p{3.0cm} X p{5.0cm} @{}}
    \toprule
    \textbf{Category} & \textbf{Fields}\\
    \midrule
    Extracted text concepts &
      Assignment; Confidentiality; Consequences of Termination;
      Dispute Resolution; Force Majeure; Indemnity;
      Limitations of Liability; Termination \\
    \addlinespace
    Duration fields &
      Term; Payment Term; Payment Period Frequency; Renewal Term;
      Renewal Notice Period; Termination Notice Window \\
    \addlinespace
    Date fields &
      Effective Date; Executed Date; End Date \\
    \addlinespace
    Currency fields &
      Annual Contract Value; Total Contract Value\\
    \addlinespace
    Boolean and categorical (closed-set selection) &
      Termination for Cause; Termination for Convenience;
      Exclusions from Liability; Renewal Type \\
    \addlinespace
    Short-text identifier fields &
      Contract Name; Parties; Governing Law \\
    \bottomrule      
    \end{tabularx}
    \caption{Field categories and constituent fields used for extraction.}
    \label{tab:fieldcat}
\end{table*}

\subsection{Data and Model Development}
\label{sec:data}

\subsubsection{Source Data}
\label{sec:sourcedata}

The model-development and evaluation data were drawn from public SEC EDGAR exhibit filings; no private, confidential, or privileged material was used. The source pool was limited to contracts between 10,000 and 100,000 tokens and to documents for which a preliminary extraction pass found at least 22 of the 26 target fields. Development data covered a diverse mix of agreement types commonly encountered in commercial contract review. 

\subsubsection{Training and Validation Split}
\label{sec:datasplit}

The training and validation splits comprised 89,517 training labels and 5,453 validation labels, stratified by contract type and separated by document identifier.

The percentage breakdown of agreement type and label prevalence in the training set can be found in Appendix~\ref{app:trainingdocs} and Appendix~\ref{app:label_prevalence}, respectively. The held-out human evaluation set is described separately in \S\ref{sec:evaldata}.

\subsubsection{Label Generation and Filtering}
\label{sec:labelgen}

Training corpus labels were generated synthetically with a frontier language model and filtered by an LLM-as-judge panel that scored answer quality and span validity. Cited spans were fuzzy-matched back to the source text. This process produced field-level labels with both display answers and supporting spans, matching the output format later required at evaluation time.

\begin{figure}[t]
\centering
\includegraphics[width=0.70\linewidth]{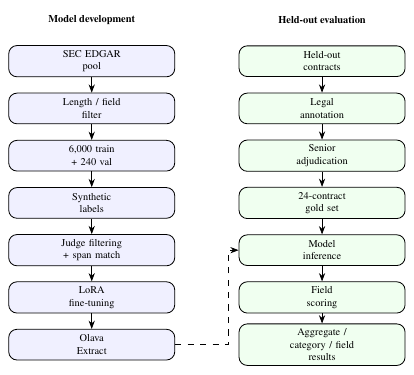}
\caption{Model-development and evaluation pipeline.}
\Description{A two-column flow diagram showing model development on the left and held-out evaluation on the right. The left column includes SEC EDGAR pool, filtering, train-validation split, synthetic labels, judge filtering, LoRA fine-tuning, and Olava Extract. The right column includes held-out contracts, legal annotation, adjudication, gold set creation, model inference, scoring, and results. A dashed arrow connects Olava Extract to model inference.}
\label{fig:pipeline}
\end{figure}

Figure~\ref{fig:pipeline} summarises the separation between the synthetically labelled model-development data and the independently human-annotated evaluation set.

\subsubsection{Model and Fine-Tuning}
\label{sec:modelFT}

 Training used parameter-efficient fine-tuning (LoRA) in bfloat16 precision and ran for one epoch over the 89,517 training labels. The final checkpoint was selected by validation loss on the 5,453 validation labels.

\subsection{Evaluation Protocol}
\label{sec:evaluation}

The evaluation protocol was designed to compare Olava Extract with frontier systems under the same task, contract set, output schema, prompt structure, inference approach, and scoring rules. This section describes the held-out evaluation set, compared models, inference procedure, field-level matching rules, accuracy metrics, and cost and latency measurement.

\subsubsection{Evaluation Dataset and Human Annotation}
\label{sec:evaldata}

The evaluation set contained 24 held-out SEC EDGAR contracts and 508 human-labelled annotations. Legal professionals annotated the contracts in Label Studio. Each contract was reviewed by at least two lawyers, with disagreements resolved by a senior lawyer. The evaluation contracts were excluded from the training and validation splits by document identifier.

\subsubsection{Compared Models}
\label{sec:models}

Olava Extract was compared with five commercially available frontier models: Claude Opus 4.6, Claude Sonnet 4.6, Gemini 2.5 Pro, Gemini 3.1 Pro Preview, and GPT-5.4. Each model was evaluated on the same 24-contract dataset using the same structured output specification. The frontier models were evaluated zero-shot: the system prompt specified the expected output schema and format but included no worked extraction examples.

\subsubsection{Inference Protocol}
\label{sec:inference}

All models received the same contracts, system prompts, user prompts, output instructions, and field requests. Consistent with the task definition in \S\ref{sec:task}, full contracts were inserted into the user message without chunking, and all 26 fields were extracted in a single model call per contract. Each contract was processed in a single run with no result averaging. No lower-casing, date normalisation, currency parsing, or LLM-judge rescoring was done at the time of evaluation.  

For Olava Extract, evaluation inference was run on two H200 SXM GPUs in bfloat16 precision without quantisation.  The model was served via vLLM 0.19.1 with prefix caching enabled, a maximum context length of 262,144 tokens, and the LoRA adapter loaded at rank 32.

\subsubsection{Field-Level Matching Rules}
\label{sec:fieldmatch}

Model outputs were scored against the human-labelled annotations using data-type-specific matching rules. The matching rules were applied to the display answer, supporting span(s), or both, depending on the field category.  Table~\ref{tab:fields} summarises the matching rule for each category.

\begin{table*}[ht]
  \centering  
  \small
  \begin{tabularx}{\textwidth}{@{} p{3.0cm} X p{5.0cm} @{}}
    \toprule
    \textbf{Category} & \textbf{Matching rule} \\
    \midrule
    Extracted text concepts &
      Reference span overlap \\
    \addlinespace
    Duration fields &
      Display-answer substring containment or reference span overlap \\
    \addlinespace
    Date fields &
      Display-answer substring containment or reference span overlap;
      no date normalisation \\
    \addlinespace
    Currency fields &
      Display-answer substring containment or reference span overlap;
      no currency normalisation \\
    \addlinespace
    Boolean and categorical (closed-set selection) &
      Exact string equality against a fixed option list \\
    \addlinespace
    Short-text identifier fields &
      Exact string equality against a string extracted from the contract \\
    \bottomrule      
    \end{tabularx}
    \caption{Field categories and matching rules used for evaluation.}
    \label{tab:fields}
\end{table*}

\subsubsection{Accuracy Metrics}
\label{sec:metrics}

Performance is reported using precision, recall, and F1. Scores were computed separately for each field and then aggregated across fields. Macro-averaged scores treat each of the 26 fields equally by averaging field-level precision, recall, and F1. Micro-averaged scores pool labelled field instances across all fields before computing precision, recall, and F1.

Both macro- and micro-averaged results are reported because the field distribution is uneven: common fields such as termination provisions and executed dates appear in most contracts, while fields such as Renewal Notice Period and Total Contract Value are less prevalent. No statistical significance testing or confidence intervals are computed.

\subsubsection{Cost and Latency Measurement}
\label{sec:costlatency}

Cost and latency were measured at the document level for each model. For Olava Extract, GPU time was priced at \$4.01 per H200-hour. Two self-hosted cost figures are computed. The parallel-batched estimate uses wall-clock time for the full 24-document batch. The unbatched serial estimate sums per-document runtimes, approximating the cost of processing documents one at a time. 

Frontier-model costs are computed from observed token usage at published API rates. The reported figures should be interpreted as an operational comparison under the tested conditions, not as a universal claim about all possible hosting configurations, hardware choices, batching strategies, provider rate limits, batching behaviour, or provider pricing structures.

\section{Results}
\label{sec:results}

This section reports the performance of Olava Extract against the five frontier baselines described in \S\ref{sec:models}. Aggregate scores are reported first, followed by a breakdown by field category and field. Unless noted, all numbers refer to F1 computed on the 24-contract human-annotated evaluation set, with matching criteria as defined in \S\ref{sec:fieldmatch}.

\subsection{Aggregate Performance}
\label{sec:aggregate}

Table~\ref{tab:micro} reports micro-averaged precision, recall, and F1 across all 26 fields. Table~\ref{tab:macro} reports the macro-averaged equivalents. Olava Extract achieves the highest scores on both aggregations, though the margin over the strongest frontier baselines is narrow.

\begin{table}[ht]
  \centering
  \begin{tabular}{@{}lccc@{}}
    \toprule
    \textbf{Model} & \textbf{Prec.} & \textbf{Rec.} & \textbf{F1} \\
    \midrule
    Olava Extract          & 0.812 & 0.874 & 0.842 \\
    Gemini 3.1 Pro Preview & 0.783 & 0.860 & 0.820 \\
    Claude Opus 4.6        & 0.777 & 0.862 & 0.817 \\
    Claude Sonnet 4.6      & 0.745 & 0.855 & 0.796 \\
    GPT-5.4                & 0.686 & 0.902 & 0.779 \\
    Gemini 2.5 Pro         & 0.693 & 0.898 & 0.783 \\
    \bottomrule
  \end{tabular}
    \caption{Micro-averaged performance across all 26 fields, ranked by F1.}
    \label{tab:micro}
\end{table}

\begin{table}[ht]
  \centering  
  \begin{tabular}{@{}lccc@{}}
    \toprule
    \textbf{Model} & \textbf{Prec.} & \textbf{Rec.} & \textbf{F1} \\
    \midrule
    Olava Extract          & 0.780 & 0.859 & 0.812 \\
    Gemini 3.1 Pro Preview & 0.764 & 0.849 & 0.796 \\
    Claude Opus 4.6        & 0.759 & 0.850 & 0.794 \\
    GPT-5.4                & 0.704 & 0.877 & 0.769 \\
    Gemini 2.5 Pro         & 0.704 & 0.878 & 0.769 \\
    Claude Sonnet 4.6      & 0.717 & 0.839 & 0.766 \\
    \bottomrule
  \end{tabular}
    \caption{Macro-averaged performance across all 26 fields, ranked by F1.}
    \label{tab:macro}
\end{table}

\begin{figure}[t]
\centering
\includegraphics[width=0.85\linewidth]{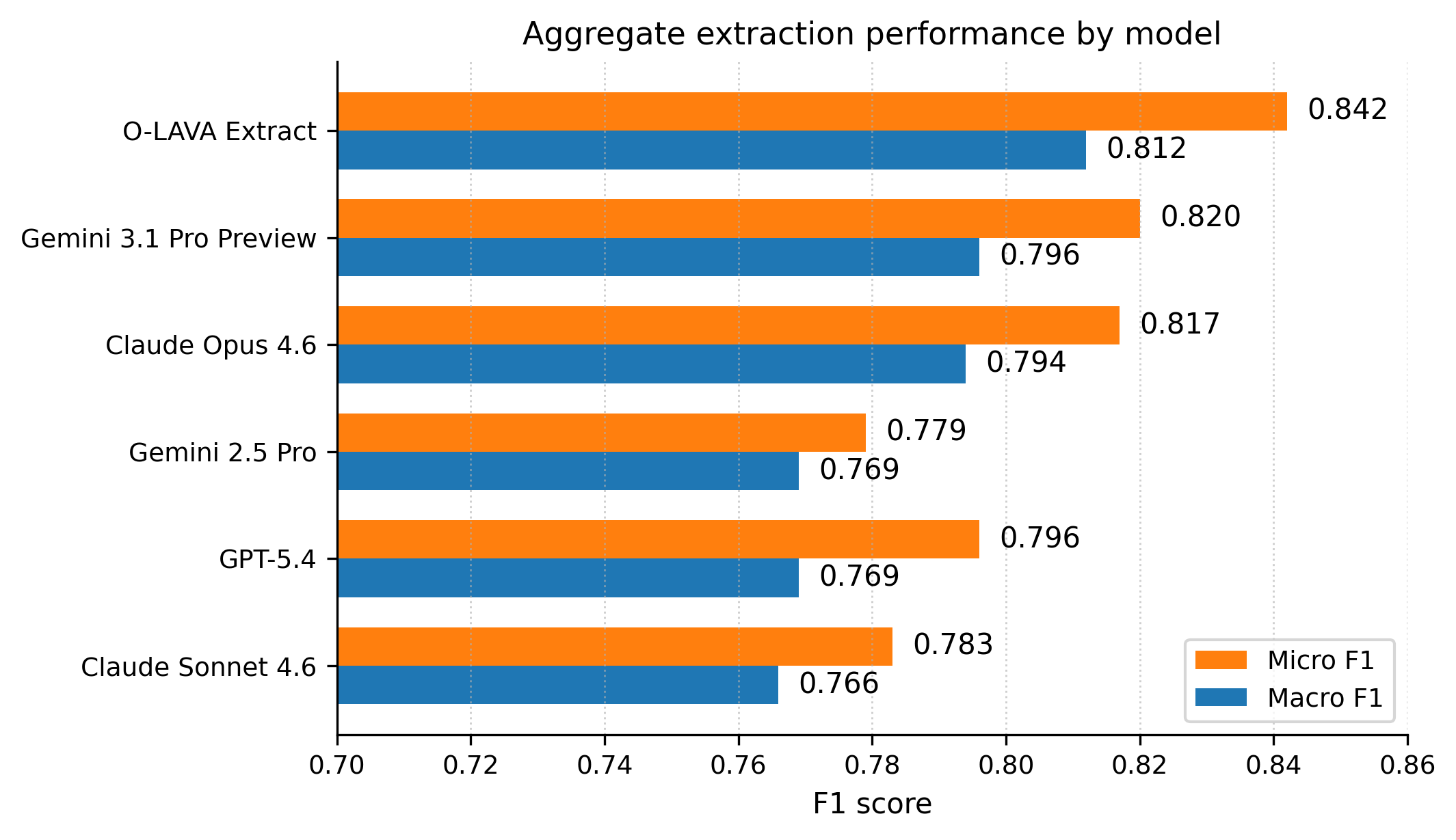}
\caption{Aggregate macro- and micro-averaged F1 across the 26-field extraction task. Olava Extract ranks highest on both measures, but the margin over the strongest frontier baseline is small.}
\Description{Horizontal grouped bar chart comparing macro F1 and micro F1 across Olava Extract, Gemini 3.1 Pro Preview, Claude Opus 4.6, GPT-5.4, Gemini 2.5 Pro, and Claude Sonnet 4.6. Olava Extract has the highest macro and micro F1, but the top models are closely clustered.}
\label{fig:aggregate-f1}
\end{figure}

Figure~\ref{fig:aggregate-f1} visualises the aggregate macro- and micro-averaged F1 scores, showing that the leading models are closely clustered.

Two observations stand out. First, Olava Extract not only matches, but leads frontier models on aggregate performance. Second, Olava Extract achieves the highest precision of any model evaluated on both aggregations (0.812 micro, 0.780 macro). This precision-recall profile meant that Olava Extract was especially strong at avoiding unsupported extractions, resulting in fewer hallucinated answers.

\subsection{Performance by Field Category}
\label{sec:bycategory}

\begin{figure}[t]
\centering
\includegraphics[width=0.85\linewidth]{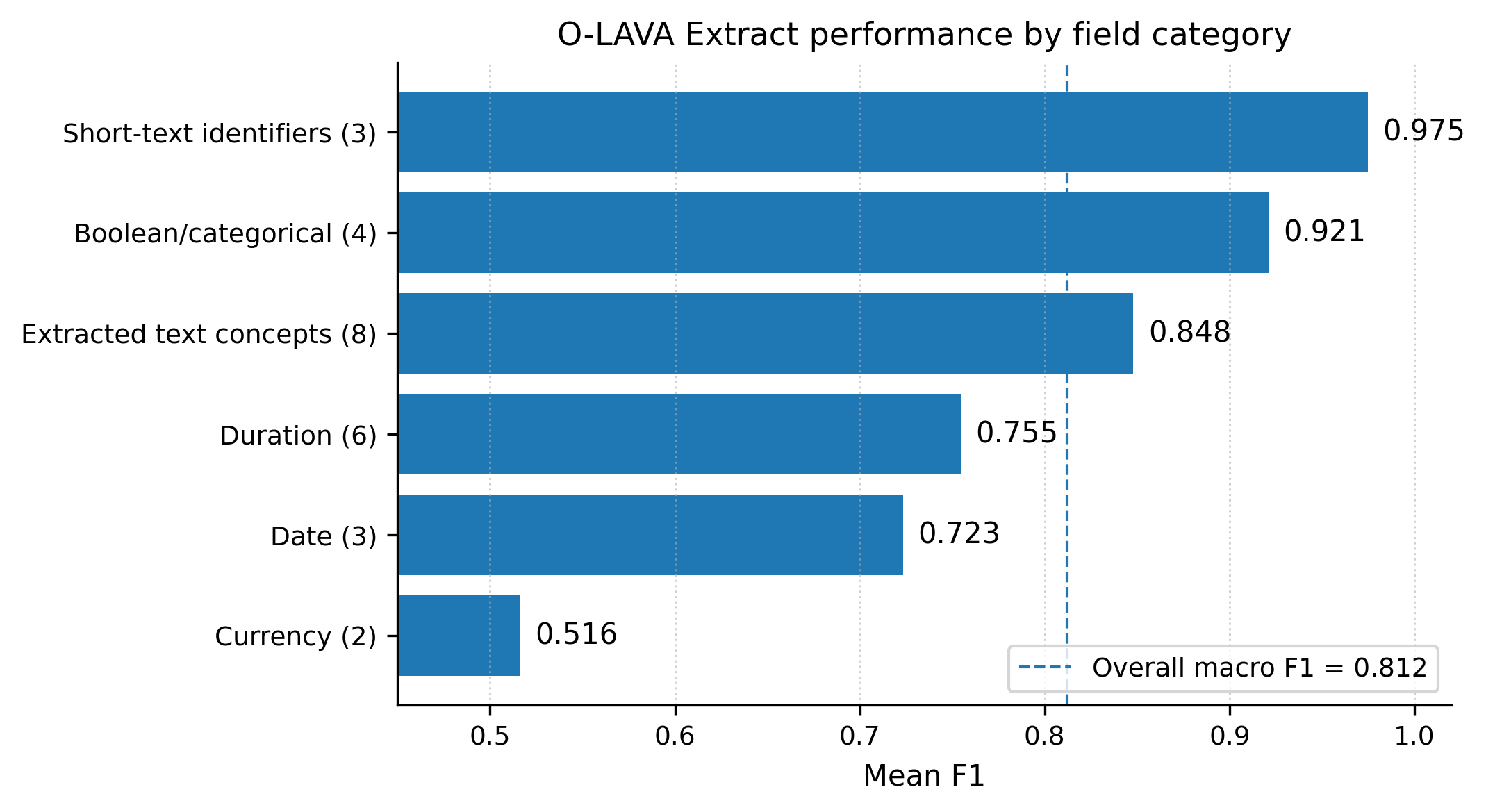}
\caption{Olava Extract mean F1 by field category. Performance is strongest on short-text identifiers and extracted text concepts, and weakest on currency fields requiring normalisation or aggregation.}
\Description{Horizontal bar chart showing Olava Extract's mean F1 by field category. Short-text identifiers have the highest mean F1, followed by extracted text concepts, Boolean and categorical fields, duration fields, date fields, and currency fields. A vertical reference line marks the model's overall macro F1.}
\label{fig:category-f1}
\end{figure}

Drilling down from the aggregate scores shows variation in model performance across the six field categories defined in \S\ref{sec:sourcedata} and scored using the matching rules in \S\ref{sec:fieldmatch}. Figure~\ref{fig:category-f1} summarises Olava Extract's mean F1 by field category. The following paragraphs present the corresponding field-level results, with the full per-field, per-model F1 grid reported in Appendix \ref{app:perfield}.

\paragraph{Extracted text concepts (8 fields).} Olava Extract performs strongly across this category, with F1 ranging from 0.787 (Consequences of Termination) to 0.920 (Assignment). It records the highest F1 of any model tested on Termination (0.887), Force Majeure (0.842), and Dispute Resolution (0.789), and comes within a small margin of the leader on Assignment (0.920 vs.\ 0.958), Limitations of Liability (0.914 vs.\ 0.933), Indemnity (0.845 vs.\ 0.875), and Confidentiality (0.800 vs.\ 0.844). The category mean is 0.848, materially above the model's overall macro F1.

\paragraph{Boolean and categorical fields (4 fields).} Performance is bimodal. Olava Extract achieves perfect F1 on Termination for Convenience (1.000), ties for the field leader on Termination for Cause (0.917), scores 0.917 on Exclusions from Liability against a leader of 0.958 and reaches 0.851 on Renewal Type.

\paragraph{Short-text identifier fields (3 fields).} This is the strongest category for Olava Extract. F1 is 0.990 on Parties, 0.977 on Governing Law, and 0.958 on Contract Name. All three are at or near the top of the field among the models tested. The category mean (0.975) is materially above the model's overall macro F1 and indicates that contract-specific named-entity extraction---identifying who the parties are, what the contract is called, and which jurisdiction governs it---is well within the model's competence.

\paragraph{Duration fields (6 fields).} Results are uneven. Olava Extract leads or ties on Renewal Notice Period (1.000), and performs competitively on Payment Term (0.882), Term (0.867), and Termination Notice Window (0.750). Performance is weaker on Renewal Term (0.571) and Payment Period Frequency (0.457). These two fields are also the weakest for most frontier models, which suggests that the gap reflects underlying ambiguity in the source data---for example, contracts that describe renewal terms across multiple sentences, or that express payment periodicity informally---rather than a weakness specific to Olava Extract alone.

\paragraph{Date fields (3 fields).} Performance is mid-tier. Effective Date (0.818) and End Date (0.720) are competitive; Executed Date (0.632) is the model's weakest of the three. No model evaluated exceeds 0.737 on Executed Date, again suggesting an annotation- or source-level difficulty rather than a model-specific weakness.

\paragraph{Currency fields (2 fields).} This is the weakest category for Olava Extract and for the field as a whole. Annual Contract Value F1 is 0.333; Total Contract Value F1 is 0.700. Performance on these two fields is uniformly poor across all models tested: no model scores above 0.700 on Total Contract Value (Olava Extract ties this upper bound) or above 0.476 on Annual Contract Value.

\subsection{Field-Level Variation}
\label{sec:fieldlevel}

Of the 26 fields, Olava Extract is the outright leader on 5, ties for leader on 4, falls within 0.05 F1 of the leader on 10 more, and trails the leader by more than 0.05 F1 on 7. The distribution is summarised in Table~\ref{tab:ranking}. The fields on which the model trails the leader are heavily concentrated in the Currency, Date, and option-list categories, consistent with the categorical analysis above.

\begin{table}[ht]
  \centering  
  \small
  \begin{tabularx}{\linewidth}{@{}p{2.7cm}c X@{}}
    \toprule
    \textbf{Result} & \textbf{N} & \textbf{Fields} \\
    \midrule
    Outright leader & 5 & Termination; Force Majeure; Dispute Resolution; Termination for Convenience; Parties \\
    \addlinespace
    Tied for leader & 4 & Termination for Cause; Renewal Notice Period; Governing Law; Total Contract Value \\
    \addlinespace
    Within 0.05 F1 of leader & 10 & Contract Name; Consequences of Termination; Payment Term; Termination Notice Window; Renewal Type; Limitations of Liability; Indemnity; Assignment; Exclusions from Liability; Confidentiality \\
    \addlinespace
    Trails leader by $>$0.05 F1 & 7 & Annual Contract Value; Payment Period Frequency; Renewal Term; Term; Effective Date; End Date; Executed Date \\
    \bottomrule
  \end{tabularx}
    \caption{Distribution of Olava Extract's per-field F1 against the best result of any model on each field.}
    \label{tab:ranking}
\end{table}

\subsection{Inference Cost and Latency}
\label{sec:cost}

Table \ref{tab:cost} reports per-document inference cost and average per-document latency for each model on the 24-contract evaluation set. For Olava Extract, we report two cost estimates because self-hosted GPU economics depend on batching. The batched estimate uses the wall-clock time for the full 24-contract batch multiplied by the GPU hourly rate, then divided by 24 documents. This reflects throughput-oriented production use. The serial unbatched single-document estimate instead sums the per-document latencies, multiplies that total by the GPU hourly rate, and divides by 24 documents. This approximates the cost of processing documents one at a time without batching. Frontier-model costs are computed from observed token usage at published API rates.

Olava Extract is materially cheaper per document than every frontier baseline, but its per-document latency is higher because parallel-batch inference trades tail latency for throughput; total wall-clock time for the full 24-contract batch was 6 minutes 27 seconds. 

\begin{table}[ht]
  \centering  
  \begin{tabular}{@{}lcc@{}}
    \toprule
    \textbf{Model} & \textbf{Cost/doc} & \textbf{Avg latency (s)} \\
    \midrule
    Olava Extract (batched) & \$0.018 & 313.69 \\
    Olava Extract (unbatched) & \$0.147 & 131.55 \\
    Gemini 2.5 Pro                    & \$0.149 &  99.7 \\
    Gemini 3.1 Pro Preview            & \$0.187 &  93.2 \\
    Claude Opus 4.6 (medium effort)   & \$0.258 & 100.3 \\
    GPT-5.4  (medium reasoning)       & \$0.262 & 194.3 \\
    Claude Sonnet 4.6 (medium effort) & \$0.456 & 350.8 \\
    \bottomrule
  \end{tabular}
    \caption{Per-document inference cost and average per-document latency.}
    \label{tab:cost}
\end{table}

\section{Discussion}
\label{sec:discussion}

This study addresses whether a domain-trained Small Language Model can serve as a credible alternative to frontier LLMs on structured contract extraction. Our discussion answers the three core questions that drive this research and then turns to what the implications are for the legal industry.

\subsection{Can a domain-trained Small Language Model reach aggregate extraction performance comparable to leading frontier models?} 
Olava Extract finished first by both aggregate measures, with a micro F1 of 0.842 and a macro F1 of 0.812. Its margins over the strongest frontier baseline are 0.022 micro F1 and 0.016 macro F1. Olava Extract competes head-to-head with the largest commercially hosted systems on full-document structured extraction. That places domain-trained SLMs firmly inside the conversation about production legal AI.

The field-level results support the same conclusion. Olava Extract led outright on 5 of 26 fields, tied for the lead on 4, and finished within 0.05 F1 of the leader on 10 more. Its strongest results were concentrated in common commercial review fields. Its weaker results were concentrated in a smaller set of fields where every model struggled, suggesting that field definition, normalisation, and matching criteria explain much of the remaining gap. On the fields most central to contract review, the SLM is at parity with or ahead of the frontier baselines.

\subsection{Are Small Language Models cheaper than Large Language Models, beyond the obvious advantages of self-hosting?}

The cost advantage is material. Under batched self-hosted inference, Olava Extract processed contracts at \$0.018 per document. The frontier API baselines, calculated from observed token usage and published pricing, ranged from \$0.149 to \$0.456 per document. The cheapest frontier baseline therefore cost more than eight times as much; the most expensive cost more than twenty-five times as much. Even in the unbatched serial estimate, Olava Extract remained competitive with the cheapest frontier baseline.

This changes the financial profile of high-volume contract review. Once volume justifies self-hosted deployment, cost no longer scales directly with every additional document. It becomes an infrastructure cost that legal teams can forecast, budget, and audit. For repository-scale extraction, due diligence review, and recurring contract analysis, that distinction matters.

\subsection{Are Small Language Models faster than Large Language Models?}

The speed result depends heavily on workload shape and deployment configuration. In single-document inference, Olava Extract averaged 131 seconds per document, compared with 194 seconds for GPT-5.4 and 351 seconds for Claude Sonnet 4.6, but slower than Claude Opus 4.6 and the Gemini models, which averaged approximately 93–100 seconds per document. The result is therefore not that the SLM is uniformly faster on per-document latency, but that it is broadly competitive with frontier systems while operating in a self-hosted configuration.

The more important result concerns throughput economics. In parallel-batched mode, Olava Extract processed all 24 contracts in 6 minutes and 27 seconds on a fixed two-H200 setup. As expected, batching increased per-document latency to 314 seconds while substantially improving total throughput. For many legal workflows—including repository extraction, diligence review, and recurring contract audits—overall throughput matters more than the latency of any individual document.

Performance is also shaped by infrastructure economics. Because Olava Extract is materially cheaper per document than the frontier API baselines, organisations can allocate substantially more inference capacity while remaining cost-competitive overall. In practice, throughput and latency become deployment decisions rather than fixed provider characteristics: additional GPUs, more aggressive batching, or parallel inference pipelines may increase throughput significantly while preserving a cost advantage over frontier APIs.

This study therefore does not claim that Small Language Models are intrinsically faster than frontier systems under all conditions. Frontier throughput depends on provider-side queueing, concurrency limits, and API allocation policies, while self-hosted performance depends on hardware allocation and inference optimisation choices. The practical finding is narrower but operationally significant: a domain-trained, self-hosted SLM can deliver competitive latency and strong batch throughput at substantially lower per-document cost and under infrastructure controlled by the deploying organisation.

\subsection{Implications}

The implications of these findings are profound for the legal industry. Prior work in this series showed that frontier Large Language Models could perform routine legal review tasks at or above human-reviewer baselines while operating substantially faster and cheaper than Junior Lawyers and Legal Process Outsourcers [9, 10]. This study extends that result further. It shows that high-performing, human-comparable legal AI can now be deployed at dramatically lower cost and at infrastructure-scale throughput using smaller domain-trained models. The economic pressure on Junior Lawyers and Legal Process Outsourcers therefore becomes substantially more acute. Tasks that historically depended on large pools of human reviewers may increasingly be performed by specialised systems that are faster, cheaper, more scalable, and deployable directly within enterprise infrastructure.

The implications also extend beyond legal operations to the economics of the AI industry itself. Over the last several years, extraordinary amounts of capital and infrastructure investment have been concentrated around increasingly large general-purpose models. The prevailing assumption has been that state-of-the-art enterprise AI capability would remain tightly coupled to frontier-scale training runs, hyperscale infrastructure, and centrally hosted model providers.

The results substantially challenge that assumption. Olava Extract, a domain-trained self-hosted Small Language Model, achieved the strongest aggregate extraction performance in this study while operating at a fraction of the inference cost of the frontier models. On a commercially meaningful enterprise task, a relatively compact specialised model was able to compete directly with systems backed by vastly larger training and infrastructure investments. This raises a broader question for the AI industry: whether continually increasing investment into ever-larger general-purpose models represents the most economically useful allocation of capital for enterprise AI workloads, particularly where specialised domain models may achieve comparable practical performance at materially lower cost.

Frontier LLMs demonstrated that routine contract review could be automated [9]. Domain-trained SLMs make that automation significantly feasible. When high-quality extraction can run inside organisation-controlled infrastructure at materially lower cost, the remaining barriers become organisational rather than technical. The disruption pressure on traditional legal review models therefore comes not only from improvements in AI capability, but from the rapid collapse of the economic and operational constraints that previously limited deployment at scale.

\subsection{Further Work}
\label{sec:futurework}

Several directions would refine and extend these findings. First, the evaluation set should be expanded to support confidence intervals, significance testing, and inter-annotator agreement, ideally stratified by contract type, length, and field prevalence.

Additionally, the weakest field categories would benefit from targeted training data, normalisation-aware matching rules, and revised field-prompt design. Future work should also test how much of the remaining gap on these fields can be closed without retraining, by varying system prompt design and the inference harness around the model.  

Finally, the benchmark should expanded to cover a broader range of contract types, including amendments, statements of work, order forms, addenda, exhibits, and schedules. This would better reflect the diversity, formatting variation, and document quality encountered in production legal operations.

Each of these steps would help narrow the field-level gaps and broaden the contract types over which a domain-trained SLM has been shown to compete with frontier systems.

\section{Conclusion}
\label{sec:conclusion}

This study shows that domain-trained Small Language Models are a viable and compelling alternative to frontier Large Language Models for structured contract extraction. Olava Extract demonstrates that reliable contract review does not require the largest externally hosted models. With targeted legal-domain training, smaller self-hostable models can deliver the accuracy, control, and efficiency required for production legal workflows.

The implications for legal operations are significant. Contract extraction has typically depended on human reviewers or increasingly expensive frontier-model APIs, both of which introduce practical constraints at scale. A domain-trained SLM changes that equation. It offers a path toward contract review that is not only automated, but also more predictable, auditable, and deployable within an organisation’s own infrastructure.

The evidence points to a clear conclusion: SLMs are not merely a compromise for legal AI; they are an emerging class of purpose-built tools capable of reshaping high-volume contract review. As legal teams continue to adopt AI, the advantage will increasingly belong to systems that combine strong legal performance with operational control, cost discipline, and deployment flexibility. Domain-trained SLMs are therefore positioned to play a central role in the next phase of legal AI adoption, extending the disruption previously identified for frontier LLMs into a more practical and scalable production model.

\begin{acks}
This research was conducted by Onit’s AI Labs, with special mention to the principal investigators and supporters of the research:
Nicole Lincoln, Principal Legal AI Researcher(PI); Nick Whitehouse, Chief AI Officer (Co-PI); Jaron Mar, Principal AI Engineer (Co-PI); Rivindu Perera, Senior Vice President, AI and Ethics (Co-PI); Seung Myung, Principal AI Engineer; Revanth Mohan, Principal AI Engineer; Dhavani Shah, AI Scientist; Ning Chen, AI Research Engineer; Xiaocui Zhang, Data Scientist; Sharon John, Legal AI Researcher; Chitra Gangawat, Legal AI Researcher; Kamlesh Vyas, Legal AI Researcher; Qijun Wang, Principal Legal Engineer.
\end{acks}

\bibliographystyle{ACM-Reference-Format}
\bibliography{SLMExtract}

\appendix
\section{Appendix}

\subsection{Per-Field F1 Across All Models}
\label{app:perfield}

Table~\ref{tab:perfield} reports per-field F1 for Olava Extract and the five frontier baselines across the 26 fields evaluated in this study, grouped by field category. 

\begin{table*}[t]
  \centering
  \caption{Per-field F1 across all models on the 24-contract evaluation set.}
  \label{tab:perfield}
  \footnotesize
  \begin{tabular}{@{}lcccccc@{}}
    \toprule
    \textbf{Field} & \textbf{Olava Extract} & \textbf{Gem 3.1 PP} & \textbf{Opus 4.6} & \textbf{Sonnet 4.6} & \textbf{GPT-5.4} & \textbf{Gem 2.5 Pro} \\
    \midrule
    \multicolumn{7}{@{}l}{\textit{Extracted text concepts}} \\
Assignment                  & 0.920 & 0.958 & 0.868 & 0.885 & 0.847 & 0.793 \\
Confidentiality             & 0.800 & 0.742 & 0.844 & 0.818 & 0.782 & 0.720 \\
Consequences of Termination & 0.787 & 0.765 & 0.722 & 0.795 & 0.660 & 0.602 \\
Dispute Resolution          & 0.789 & 0.757 & 0.766 & 0.744 & 0.603 & 0.667 \\
Force Majeure               & 0.842 & 0.824 & 0.824 & 0.800 & 0.737 & 0.778 \\
Indemnity                   & 0.845 & 0.806 & 0.838 & 0.875 & 0.874 & 0.875 \\
Limitations of Liability    & 0.914 & 0.933 & 0.811 & 0.750 & 0.720 & 0.684 \\
Termination                 & 0.887 & 0.795 & 0.796 & 0.771 & 0.739 & 0.755 \\
    \addlinespace
    \multicolumn{7}{@{}l}{\textit{Boolean and categorical}} \\
    Termination for Cause       & 0.917 & 0.875 & 0.917 & 0.851 & 0.875 & 0.917 \\
Termination for Convenience & 1.000 & 0.958 & 0.917 & 0.894 & 0.958 & 0.958 \\
Exclusions from Liability   & 0.917 & 0.917 & 0.958 & 0.894 & 0.917 & 0.917 \\
Renewal Type                & 0.851 & 0.870 & 0.766 & 0.652 & 0.809 & 0.723 \\
    \addlinespace
    \multicolumn{7}{@{}l}{\textit{Categorical / short-text identifier}} \\
Contract Name               & 0.958 & 0.958 & 0.792 & 0.875 & 0.962 & 0.958 \\
Parties                     & 0.990 & 0.970 & 0.980 & 0.979 & 0.970 & 0.980 \\
Governing Law               & 0.977 & 0.977 & 0.955 & 0.955 & 0.923 & 0.955 \\
    \addlinespace
    \multicolumn{7}{@{}l}{\textit{Duration}} \\
Term                        & 0.867 & 0.839 & 0.813 & 0.788 & 0.929 & 0.727 \\
Payment Term                & 0.882 & 0.800 & 0.833 & 0.833 & 0.475 & 0.895 \\
Payment Period Frequency    & 0.457 & 0.400 & 0.629 & 0.500 & 0.512 & 0.439 \\
Renewal Term                & 0.571 & 0.667 & 0.615 & 0.571 & 0.667 & 0.571 \\
Renewal Notice Period       & 1.000 & 0.750 & 1.000 & 1.000 & 1.000 & 1.000 \\
Termination Notice Window   & 0.750 & 0.688 & 0.710 & 0.606 & 0.765 & 0.611 \\
    \addlinespace
    \multicolumn{7}{@{}l}{\textit{Date}} \\
Effective Date              & 0.818 & 0.791 & 0.818 & 0.773 & 0.864 & 0.884 \\
Executed Date               & 0.632 & 0.703 & 0.615 & 0.667 & 0.667 & 0.737 \\
End Date                    & 0.720 & 0.833 & 0.692 & 0.714 & 0.783 & 0.833 \\
    \addlinespace
    \multicolumn{7}{@{}l}{\textit{Currency}} \\
Annual Contract Value       & 0.333 & 0.435 & 0.476 & 0.320 & 0.250 & 0.370 \\
Total Contract Value        & 0.700 & 0.700 & 0.700 & 0.600 & 0.700 & 0.700 \\
    \bottomrule
  \end{tabular}
\end{table*}

\subsection{Training Document Breakdown}
\label{app:trainingdocs}

Table~\ref{tab:doc_types} reports the distribution of agreement types in the model-development corpus. The training set covers a range of commercial contract categories, with no single category dominating the corpus.

\begin{table}[ht]
\centering
\begin{tabular}{lc}
\hline
\textbf{Type} & \textbf{\% of Total Documents} \\
\hline
Finance                      & 12.2\% \\
Employment Vertical          & 12.2\% \\
Shareholder Agreement        & 12.2\% \\
Procurement Review 2020      & 12.2\% \\
Property Review              & 12.2\% \\
Sales and Marketing          & 12.2\% \\
SaaS and Technology Skillset &  7.4\% \\
NDA                          &  6.8\% \\
SOW                          &  6.8\% \\
Settlement Agreement Review  &  4.0\% \\
Construction                 &  1.8\% \\
\hline
\textbf{Total}               & \textbf{100\%} \\
\hline
\end{tabular}
\caption{Document type distribution across the dataset.}
\label{tab:doc_types}
\end{table}

\subsection{Label Prevalence}
\label{app:label_prevalence}

Table~\ref{tab:label_prevalence} reports the prevalence of each extraction field in the model-development corpus. Field prevalence varies substantially, which helps contextualise the field-level results reported in Appendix \ref{app:perfield} and the weaker performance observed on some low-frequency targets.

\begin{table}[ht]
\centering
\begin{tabular}{lc}
\hline
\textbf{Field} & \textbf{\% of Documents with Label} \\
\hline
Termination                 & 98.5\% \\
Executed Date               & 97.2\% \\
Dispute Resolution          & 96.9\% \\
Termination for Cause       & 96.0\% \\
Effective Date              & 95.4\% \\
Consequences of Termination & 94.2\% \\
Confidentiality             & 94.1\% \\
Indemnity                   & 93.7\% \\
Term                        & 79.0\% \\
Payment Period Frequency    & 73.4\% \\
Payment Term                & 72.5\% \\
Force Majeure               & 65.6\% \\
Termination Notice Window   & 64.8\% \\
Renewal Term                & 46.3\% \\
End Date                    & 44.7\% \\
Annual Contract Value       & 33.1\% \\
Renewal Notice Period       & 25.1\% \\
Total Contract Value        & 23.0\% \\
\hline
\end{tabular}
\caption{Label prevalence across the dataset.}
\label{tab:label_prevalence}
\end{table}

\end{document}